\documentclass[conference]{IEEEtran}
\usepackage[T1]{fontenc}
\IEEEoverridecommandlockouts
\usepackage{cite}
\usepackage{amsmath,amssymb,amsfonts}
\usepackage{algorithmic}
\usepackage{graphicx}
\usepackage{textcomp}

\usepackage{tabularx}
\usepackage{times}  
\usepackage{helvet}  
\usepackage{courier}  
\usepackage[hyphens]{url}  
\usepackage{graphicx} 
\usepackage{caption} 
\usepackage{algorithm}
\usepackage{algorithmic}
\usepackage{amsmath}
\usepackage{amssymb}
\usepackage{array}
\usepackage{booktabs}
\usepackage{longtable}
\usepackage[utf8]{inputenc}
\usepackage{enumitem}
\usepackage{subcaption}  
\usepackage{pdfpages}

\usepackage{booktabs}
\usepackage{multirow}

\ifCLASSINFOpdf
\else
\fi
\usepackage{amsmath}
\usepackage{amssymb}
\usepackage{hyperref}
\usepackage{tabularx}
\usepackage{booktabs}
\usepackage{array}
\usepackage{adjustbox}

\usepackage[most]{tcolorbox}
\usepackage{xcolor}
\newtcolorbox{ashbox}{
  breakable,
  colback=black!6,    
  colframe=black!25,  
  boxrule=0.4pt,
  left=1.2ex,right=1.2ex,top=1ex,bottom=1ex,
  sharp corners
}

\begin{document}
%
\title{Domain-Specific Knowledge Graphs in RAG-Enhanced Healthcare LLMs}

\author{
\IEEEauthorblockN{Sydney Anuyah\textsuperscript{*}, Mehedi Mahmud Kaushik\textsuperscript{*}, Hao Dai\textsuperscript{\ddag}, Rakesh Shiradkar\textsuperscript{\dag}, Arjan Durresi\textsuperscript{*}, Sunandan Chakraborty\textsuperscript{*}}
\IEEEauthorblockA{\textsuperscript{*}Luddy School of Informatics, Computing, and Engineering, Indiana University, Indianapolis, IN, USA}
\IEEEauthorblockA{\textsuperscript{\ddag}School of Medicine, Indiana University, Indianapolis, IN, USA}
\IEEEauthorblockA{\textsuperscript{\dag}Department of Biomedical Engineering and Informatics, Indiana University, Indianapolis, IN, USA}
\IEEEauthorblockA{Emails: \{sanuyah, mekaush, daihao, rshirad, adurresi, sunchak\}@iu.edu}
}


%


\maketitle

\begin{abstract}
Large Language Models (LLMs) generate fluent answers but can struggle with trustworthy, domain-specific reasoning. We evaluate whether domain knowledge graphs (KGs) improve Retrieval-Augmented Generation (RAG) for healthcare by constructing three PubMed-derived graphs: \textbf{$\mathbb{G}_1$} (T2DM), \textbf{$\mathbb{G}_2$} (Alzheimer’s disease), and \textbf{$\mathbb{G}_3$} (AD+T2DM). We design two probes: \textbf{Probe~1} targets merged AD--T2DM knowledge, while \textbf{Probe~2} targets the intersection of \textbf{$\mathbb{G}_1$}  and  \textbf{$\mathbb{G}_2$}. Seven instruction-tuned LLMs are tested across retrieval sources \{\textit{No-RAG}, $\mathbb{G}_1$, $\mathbb{G}_2$, $\mathbb{G}_1${+}$\mathbb{G}_2$, $\mathbb{G}_3$, $\mathbb{G}_1${+}$\mathbb{G}_2${+}$\mathbb{G}_3$\} and three decoding temperatures. Results show that \emph{scope alignment} between probe and KG is decisive: precise, scope-matched retrieval (notably $\mathbb{G}_2$) yields the most consistent gains, whereas indiscriminate graph unions often introduce distractors that reduce accuracy. Larger models frequently match or exceed KG-RAG with a \textit{No-RAG} baseline on Probe~1, indicating strong parametric priors, whereas smaller/mid-sized models benefit more from well-scoped retrieval. Temperature plays a secondary role; higher values rarely help. We conclude that \textbf{precision-first, scope-matched KG-RAG} is preferable to breadth-first unions, and we outline practical guidelines for graph selection, model sizing, and retrieval/reranking. \href{https://github.com/sydneyanuyah/RAGComparison}{Code and Data available here - https://github.com/sydneyanuyah/RAGComparison}
\end{abstract}


%
\IEEEpeerreviewmaketitle

\section{Introduction}
Alzheimer's disease (AD) and type 2 diabetes mellitus (T2DM) represent two of the most pressing chronic health challenges we face today \cite{vb2025global, lopez2024recent}, each carrying its own significant public health burden while also sharing surprising connections with one another \cite{a2014linking}. Today, AD stands as the primary cause of dementia in older adults \cite{alzheimer20152015, mezey2000advance}. More than 55 million people worldwide are living with dementia, and experts predict this could surge to somewhere between 139 and 150 million by 2050 \cite{schwarzinger2022forecasting, logroscino2020prevention}. T2DM shows a similarly alarming trend, with roughly 589 million adults as of 2024 (that's about 11.1\% of the adult population) living with the condition, and projections suggest we could see around 853 million cases by mid-century \cite{kalyani2025diagnosis}. What is particularly striking is that over 90\% of all diabetes cases are type 2 \cite{kalyani2025diagnosis, zheng2018global}, largely driven by our ageing populations and modern lifestyle factors. As life expectancy continues to climb, there are statistically more people dealing with both metabolic and neurodegenerative diseases simultaneously \cite{procaccini2016role}, which makes understanding the relationship between AD and T2DM increasingly urgent for biomedical AI development.

 Epidemiological research has consistently shown that having diabetes substantially increases the risk of cognitive decline and dementia, including AD \cite{barbagallo2014type}. A recent 2024 meta-analysis showed that diabetic patients face about 59\% higher dementia risk compared to people without diabetes\cite{barbagallo2014type, bellia2022diabetes}, which reinforces what earlier studies had found that T2DM increases an individual's risk of cognitive disorders by roughly 1.3 to 1.9 times \cite{kciuk2024alzheimer}. When we dig into the molecular details, the picture gets even more interesting. The chronic high blood sugar and insulin resistance that defines T2DM actually mirrors some of the same pathophysiological processes we see in AD \cite{blazquez2014insulin}. This similarity has led some researchers to provocatively label AD as ``type 3 diabetes" though this remains controversial \cite{abdelgadir2017effect, de2014type, kroner2009relationship, leszek2017type}. The idea emphasizes the overlapping features: impaired insulin signaling in the brain, chronic inflammation, and oxidative stress \cite{kciuk2024alzheimer}. Both conditions also share common risk factors such as midlife obesity and high blood pressure \cite{kciuk2024alzheimer}.

Despite the wealth of published research on AD–T2DM connections, Large Language Models (LLMs) on their own have real trouble delivering reliable medical answers \cite{lacerdaevaluation}. Today's state-of-the-art (SOTA) LLMs can certainly generate impressively fluent responses, but they frequently hallucinate facts or fabricate citations that do not exist, which is a serious safety concern in healthcare settings \cite{wang2025trustworthy}. Even powerful models like GPT-4 have inherent knowledge limitations that mean they might miss crucial domain-specific details or misrepresent the latest findings. This is where Retrieval-Augmented Generation (RAG) has become the go-to solution, helping ground LLM outputs in external, verified knowledge \cite{wu2025multirag}. By pulling in relevant documents or facts when answering a query, RAG significantly improves factual accuracy \cite{wu2025multirag}. However, conventional RAG systems that retrieve free-text passages can still drag in irrelevant or contradictory information, which the LLM might then mistakenly weave into its answer \cite{wu2025multirag}. This becomes especially problematic for knowledge-intensive questions that require multiple reasoning steps, where models need to piece together complex biomedical relationships. 

\begin{table}[ht]
\centering
\caption{Examples of triples used for the knowledge graphs}
\scriptsize
\setlength{\tabcolsep}{4pt}
\renewcommand{\arraystretch}{1.1}
\resizebox{\linewidth}{!}{%
\begin{tabular}{p{.16\linewidth} p{.84\linewidth}}
\hline
\textbf{KG} & \textbf{Example} \\
\hline
\textbf{$\mathbb{G}_1$} & [{"Entity 1": "T2DM", "Relationship": "was associated with", "Entity 2": "decreased forced expiratory volume in 1s (FEV1)"}] \\
\textbf{$\mathbb{G}_2$} & [{"Entity 1": "Alzheimer's disease CSF", "Relationship": "is associated with", "Entity 2": "neuroinflammation"}] \\
\textbf{$\mathbb{G}_3$} & [{"Entity 1": “Insulin-like growth factor ", "Relationship": "influences", "Entity 2": “cognitive functions"}] \\
\hline
\end{tabular}}
\label{tab:kg_examples}
\end{table}

Structured knowledge bases offer a more promising alternative. Knowledge graphs (KGs) organize facts as subject–relation–object triples $(E_1, R, E_2)$ as seen in Table~\ref{tab:kg_examples}, where $E_1$ is the subject entity, $E_2$ is the object entity, and $R$ represents their relationship. This structured representation provides clear semantics and traceable origins, making it easier to verify how an answer was constructed. Yet, critical questions remain about how best to leverage KGs in healthcare AI: Does curating domain-specific KGs from biomedical literature actually improve answer quality in RAG-enhanced LLMs? Should we build narrow, disease-focused graphs or broader cross-domain ones? And how do these design choices interact with model parameters like decoding temperature?

In this research, we investigate the role of domain-specific KGs curated from PubMed abstracts in RAG-enhanced healthcare LLMs. We ask whether such KGs improve the factual accuracy and evidential support of answers compared to no-RAG baselines (\textbf{RQ1}). We further examine how curation scope affects question answering: do focused, disease-specific KGs (G1 for T2DM, G2 for AD) outperform a broader merged KG (G3) on single-hop and multi-hop probes, or vice versa (\textbf{RQ2})? We also investigate the robustness of RAG gains across different decoding temperatures (0, 0.2, 0.5) and test for statistical significance of improvements (\textbf{RQ3}). Finally, we explore whether we can enhancing the components of the CoDe-KG pipeline to yield higher-quality graphs that translate into better downstream QA performance (\textbf{RQ4}).

%

\subsection{Contributions}
In this research, we investigate the role of domain-specific KGs curated from abstracts on PubMed for RAG-enhanced healthcare LLMs. 
\begin{itemize}
    \item We build three abstract-derived KGs with a common schema: \textbf{$\mathbb{G}_1$} (T2DM-focused), \textbf{$\mathbb{G}_2$} (AD-focused), and \textbf{$\mathbb{G}_3$} (combined AD+T2DM).
   \item We introduce probe sets spanning single-hop clinical facts (e.g., drug$\rightarrow$outcome, gene$\rightarrow$disease) and multi-hop mechanisms (e.g., insulin signaling$\rightarrow$neuroinflammation$\rightarrow$cognition)
   \item We evaluate 7 LLMs \emph{with} and \emph{without} KG-RAG across varying temperatures $0$, $0.2$, and $0.5$, measuring F$_1$score, and accuracy across the probes, and we test for significance of RAG gains and of curation scope ($\mathbb{G}_1$/$\mathbb{G}_2$ vs.\ $\mathbb{G}_3$)
   \item We improve the co-reference of the CoDe-KG pipeline.
\end{itemize}




\section{Background}
\subsection{Pathophysiological Links Between AD and T2DM}
AD and T2DM share several disease mechanisms that play out at both the molecular and whole-body at scales. Chronic insulin resistance, the defining feature of T2DM, contributes to AD by interfering with how neurons take up glucose and respond to insulin signals in the brain \cite{kciuk2024alzheimer}. This neurodegenerative change: the combination of excess glucose and disrupted insulin signaling fuels inflammation and oxidative stress, which then accelerates the formation of $\beta$-amyloid plaques and tau tangles (the signature brain abnormalities we see in AD) \cite{kciuk2024alzheimer}.

As established by the KG, Apolipoprotein $\epsilon$4 (APOE $\epsilon$4) offers a genetic link between these two diseases, example of a triple that shows this link is "Individuals with T2DM have APOE4-related cognitive and olfactory impairment". While we know APOE $\epsilon$4 as the strongest genetic risk factor for late-onset AD, its influence extends beyond just amyloid processing; it's also tied to how our bodies handle fats and glucose \cite{yassine2020apoe}. People with diabetes who carry APOE $\epsilon$4 experience faster mental decline and face higher dementia risk than diabetics without this gene variant, suggesting that the allele and metabolic stress amplifies each other's effects \cite{kciuk2024alzheimer, jabeen2022genetic}.

Inflammation represents another critical connection point between T2DM and AD. In T2DM, visceral fat and insulin-resistant tissues pump out elevated levels of adipokines and inflammatory cytokines. These molecules can breach the blood-brain barrier, adding fuel to the neuroinflammation already present in AD. From a clinical standpoint, T2DM patients carry roughly twice the risk of developing vascular dementia and show higher rates of Alzheimer's dementia across many studies \cite{kciuk2024alzheimer, exalto2012update}.

\subsection{Knowledge Graphs in Biomedical Research}
The complex web of relationships in biomedicine has pushed researchers to develop knowledge graphs (KGs) as a way to organize factual information in a structured, queryable format. Unlike traditional text documents, KGs represent biomedical entities like genes, diseases, drugs, proteins as nodes or entities in a network, with their relationships depicted as labelled edges. For example, an edge might connect a gene node to a disease node with the label ``associated with," making the relationship explicit and machine-readable. Large-scale biomedical KGs like SPOKE show how they created KGs by integrating over 40 curated databases, including DrugBank and GWAS catalogs, resulting in a comprehensive network of approximately 42 million nodes across 28 entity types and 160 million relationships \cite{soman2024biomedical}. CoDe-KG \cite{anuyah2025automated} is another SOTA pipeline built with open-source LLMs. What sets KGs apart from text-based resources is their built-in traceability. While bioinformaticians have traditionally used KGs for drug repurposing and gene discovery, we leverage them as knowledge sources for question answering (QA).

Given this context, we propose to leverage CoDe-KG \cite{anuyah2025automated}, an automated KG construction pipeline, to build focused biomedical knowledge graphs and evaluate their impact on trustworthy domain-specific questions. CoDe-KG is an open-source framework that extracts structured facts from text by combining robust co-reference resolution with syntactic decomposition. This approach breaks down complex statements and resolves pronouns, thereby capturing more complete and context-rich relations. According to the authors, CoDe-KG achieved an increase of $\Delta = $ 8\% F$_1$  when compared to prior methods on the REBEL relation extraction dataset \cite{anuyah2025automated}, and integrating co-reference + decomposition increased recall on rare relations by over 20\% \cite{ anuyah2025automated}. We harness this pipeline to distil knowledge from biomedical abstracts into three KGs of varying scope: $\mathbb{G}_1, \mathbb{G}_2$ and $\mathbb{G}_3$ for the combined AD+T2DM domain. We then use these graphs in an LLM-driven RAG setup to answer questions. By comparing performance across $\mathbb{G}_1, \mathbb{G}_2$, and $\mathbb{G}_3$, we examine whether a targeted disease-specific KG yields better answers than a broader cross-domain KG, or vice versa. We further probe how the structure and scope of knowledge graphs affect the LLM’s ability to deliver correct, well-supported answers. While RAG models often outperform non-RAG approaches, this is not always guaranteed. As \cite{gao2025frag, linders2025knowledge} point out, earlier KG-RAG frameworks with fixed search parameters could retrieve redundant trivial facts or miss important multi-hop connections, ultimately weakening the LLM's reasoning capabilities and making RAG-enhanced systems perform worse than their non-RAG counterparts.
\subsection{Our Approach}
In this work, we built three different knowledge graphs ($\mathbb{G}_1, \mathbb{G}_2$ and $\mathbb{G}_3$) to investigate how the scope of knowledge affects QA performance in this domain. We detail the curation of Probe1 and Probe2 in the subsequent sections. 
In Probe1, formulated on $\mathbb{G}_3$ and Probe2, formulated on $\mathbb{G}_1 \cap \mathbb{G}_2$, we tested different combinations of LLMs built on the different RAG systems.
Our hypothesis going in was that the focused KG ($\mathbb{G}_1$ and $\mathbb{G}_2$) might perform better on questions specifically about the AD-T2DM intersection, since it provides denser, more concentrated context for that particular overlap. Meanwhile, the merged KG ($\mathbb{G}_3$) should theoretically handle a wider range of questions about either disease or their connections more effectively. However, there's a potential downside to the merged approach as it could introduce distractors (facts relevant to one disease but not the question at hand), which might confuse the LLM. By testing the same set of questions against both knowledge graphs, we can observe any trade-offs between having a highly focused knowledge source versus a more comprehensive one.
\section{Methodology}
\label{sec:methodology}
\begin{figure*}[htbp]
 \centering
 \includegraphics[width=1\textwidth]{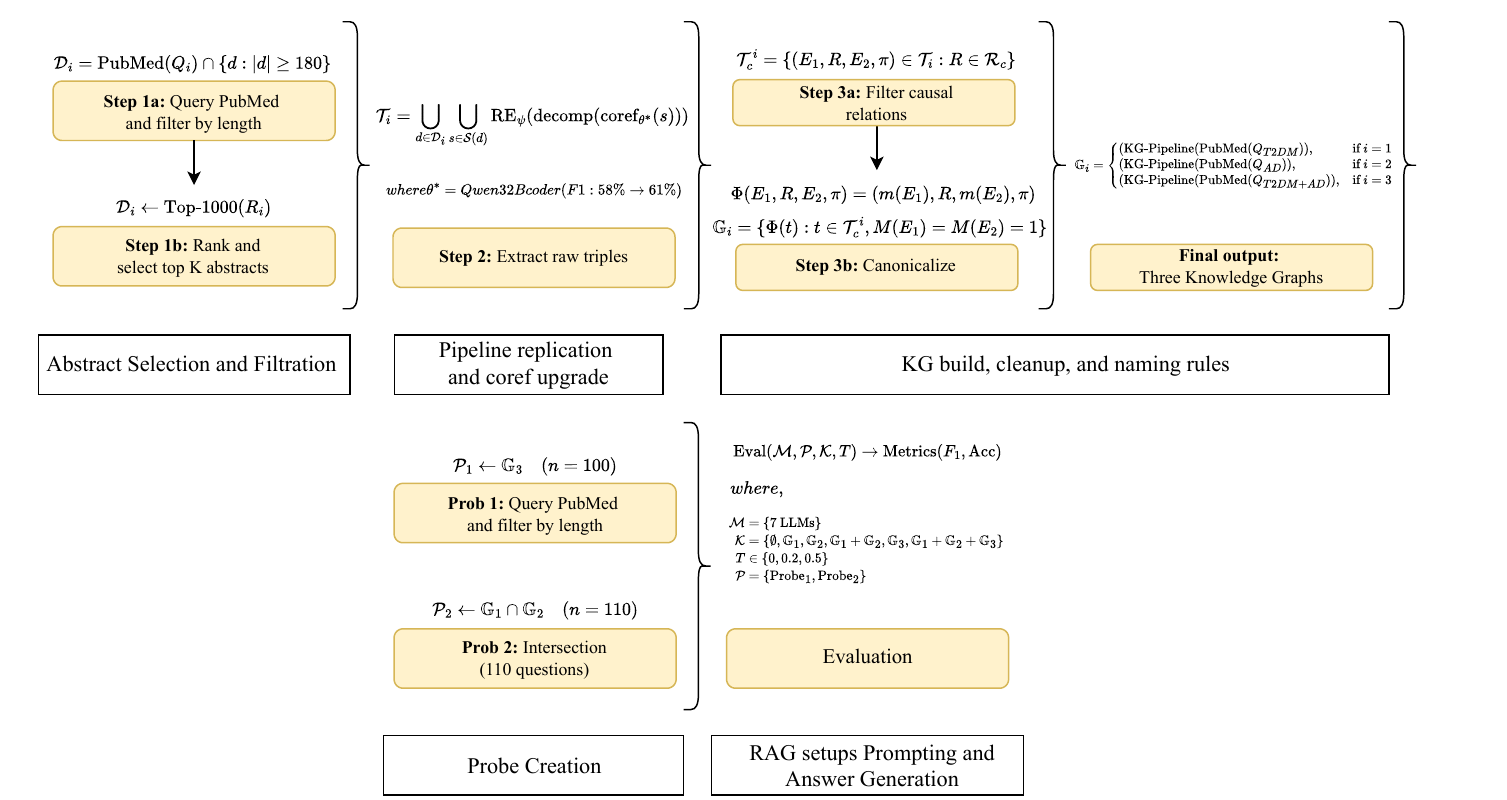}
 \caption{Overview of the methodology for constructing domain-specific knowledge graphs and evaluating their impact on RAG-enhanced healthcare LLMs. The pipeline includes abstract selection, knowledge graph construction, probe generation, and systematic evaluation across multiple models and retrieval configurations. Detailed explanation is provided in Section \ref{sec:methodology}.}
\label{fig:framework}
\vspace{-3mm}
\end{figure*}

In this section, we cover the development of the three KGs, (\textbf{$\mathbb{G}_1$} (T2DM-focused), \textbf{$\mathbb{G}_2$} (AD-focused), and \textbf{$\mathbb{G}_3$} (combined AD+T2DM)), the creation of the probes and the experimentation of the seven LLMs with different combinations of the knowledge graphs, shown in Figure \ref{fig:framework}. 

\subsection{Abstract Selection  and Filtration}

\begin{ashbox}
\textbf{Search queries used to build the KG sets} 

\vspace{0.4em}
\noindent\textbf{$Q_{T2DM}$ (T2DM-only).}\\
\texttt{(T2DM OR "Type 2 Diabetes" OR "Type II Diabetes" OR "Type-2 Diabetes" OR "Diabetes Mellitus, Type 2" OR NIDDM OR "non insulin dependent diabetes" OR "non-insulin-dependent diabetes" OR "adult-onset diabetes" OR (diabet* AND ("type 2" OR T2DM)))}

\vspace{0.4em}
\noindent\textbf{$Q_{AD}$ (AD-only).}\\
\texttt{(AD OR "Alzheimer's disease" OR "Alzheimer disease" OR "Alzheimers disease" OR Alzheimers OR Alzheime* OR Alzhiemer* OR "dementia of the Alzheimer type" OR DAT OR LOAD OR "late-onset Alzheimer*")}

\vspace{0.4em}
\noindent\textbf{$Q_{T2DM + AD}$ (AD \& T2DM together).}\\
\texttt{((AD OR "Alzheimer* disease" OR Alzheime* OR Alzhiemer* OR DAT OR LOAD) AND (T2DM OR "Type 2 Diabetes" OR "Type II Diabetes" OR "Diabetes Mellitus, Type 2" OR NIDDM))}

\end{ashbox}

After applying the $Q_{T2DM}$/$Q_{AD}$/$Q_{AD + T2DM}$ search strings, we created a simple, reproducible filter to rank abstracts and kept the most relevant \emph{per group}. The aim is to bias toward causal and mechanistic content while still covering phenotypes and biomarkers that matter for AD, T2DM, and their intersection. We remove short-worded abstracts (less than 180 words) to avoid editorials or thin notes:
\[
\text{keep}(i)\;=\;\mathbb{1}\big[\mathrm{words}(\texttt{Abstract}_i)\ge 180\big].
\]
We then join title and abstract into one text string $x_i$, and build a TF--IDF matrix on unigrams and bigrams (English stopwords, \texttt{min\_df}=2):
\[
X \in \mathbb{R}^{n \times V},\qquad X_i=\mathrm{tfidf}(x_i),\quad i=1\ldots n.
\]
We create three query vectors by TF--IDF, transforming the curated term lists:
\[
q_{\mathrm{caus}},\ q_{\mathrm{pheno}},\ q_{\mathrm{biom}} \;\in\; \mathbb{R}^V,
\]
representing  \texttt{CAUSALITY\_TERMS}, \texttt{PHENOTYPE\_TERMS}, and \texttt{BIOMARKER\_TERMS} respectively. For each abstract, we compute three cosine similarities:
\[
s^{(\mathrm{caus})}_i=\cos(X_i,q_{\mathrm{caus}}),\quad
s^{(\mathrm{pheno})}_i=\cos(X_i,q_{\mathrm{pheno}}),\quad \]
\[
s^{(\mathrm{biom})}_i=\cos(X_i,q_{\mathrm{biom}}).
\]

We reward abstracts that explicitly name crucial elements (e.g., “Mendelian randomization,” “longitudinal,” “p-tau-217,” “APOE4,” “insulin resistance phenotype,” “HbA1c”).
\[
k^{(\mathrm{caus})}_i,\ k^{(\mathrm{pheno})}_i,\ k^{(\mathrm{biom})}_i \in \mathbb{N},\qquad
k^{(\mathrm{tot})}_i = k^{(\mathrm{caus})}_i + k^{(\mathrm{pheno})}_i + k^{(\mathrm{biom})}_i.
\]

\paragraph*{Per-feature normalization.}
Each signal is min--max normalized to $[0,1]$ to keep ranges comparable:
\[
\tilde{s}^{(\cdot)}_i=\frac{s^{(\cdot)}_i-\min s^{(\cdot)}}{\max s^{(\cdot)}-\min s^{(\cdot)}},\quad
\tilde{k}_i=\frac{k^{(\mathrm{tot})}_i-\min k^{(\mathrm{tot})}}{\max k^{(\mathrm{tot})}-\min k^{(\mathrm{tot})}}.
\]
We then produce a single value that we use \emph{only to rank} abstracts:
\[
R_i \;=\; w_{\mathrm{caus}}\tilde{s}^{(\mathrm{caus})}_i
\;+\; w_{\mathrm{pheno}}\tilde{s}^{(\mathrm{pheno})}_i
\;+\; w_{\mathrm{biom}}\tilde{s}^{(\mathrm{biom})}_i \]
\[
\;+\; w_{\mathrm{kw}}\tilde{k}_i.\]

The final KGs were created from the top 1000 selected abstracts. 

\subsection{Pipeline replication and coref upgrade}
We start from the selected abstracts $\mathcal{D}=\{d_1,\ldots,d_N\}$ and treat each abstract as a set of sentences $\mathcal{S}(d)=\{s_1,\ldots,s_{|d|}\}$. The extraction loop in \textsc{CoDe-KG} is
\begin{equation}
\label{eq:pipeline}
\mathcal{T} \;=\; \bigcup_{d\in\mathcal{D}} \;\; \bigcup_{s\in \mathcal{S}(d)} \; \mathrm{RE}_\psi\!\left(\mathrm{decomp}\!\left(\mathrm{coref}_\theta(s)\right)\right),
\end{equation}
where  $s$ (\textit{coref}) $\rightarrow$ is resolving abstract co-reference, i.e ``T2DM" = ``Type 2 Diabetes Mellitus" = ``t2dm"; then decompose complex sentences into simpler ones (\textit{decomp}), then extract triples $(E_1, R, E_2)$ with a simple source tag $\pi(t)$ that notes (paper id, sentence id, clause id). This tag lets us always point back to the exact sentence that created a triple. We first matched the original authors’ inference setting (temp $=0.7$) and reproduced similar relation extraction behavior. We replaced the coreference backbone with {Qwen~32B coder} and kept all other steps the same.
\[
\mathrm{coref}_{\theta^\star} \text{ reaches } \mathrm{F1}=61\% \text{ vs. } 58\% \text{ before,}
\]
This test was done on the authors dataset. In practice, better coref means fewer broken heads or tails later, so the KG has cleaner nodes and more usable causal links.

\subsection{KG build, cleanup, and naming rules}
Raw extraction gives a multi-set of triples $\mathcal{T}$. For our study, we want edges that point in a clear causal direction, so we kept only a small relation set. 
\[
\mathcal{R}_c=\{\texttt{causes},\ \texttt{because}, ...\}.
\]
Formally,
\begin{equation}
\label{eq:causal-filter}
\mathcal{T}_c \;=\; \{\, (E_1, R, E_2,\pi) \in \mathcal{T} \;:\; r\in \mathcal{R}_c \,\}.
\end{equation}
Next, we remove vague heads/tails. A simple mask $M$ drops items like ``it'', ``this'', or ``this study'':
\[
M(x)=\mathbb{1}\!\left[x \notin \{\texttt{it},\texttt{this},\texttt{this study}\} \right],
\]
\[
\qquad
\mathcal{T}_{c}^{\mathrm{clean}}=\{t\in\mathcal{T}_c: M(E_1)=M(E_2)=1\}.
\]

We manually normalize names so variants collapse to a single label: ``T2DM'' and ``Type~2~Diabetes'' should be treated as one thing. We rewrite each edge by
\[
\Phi:\ (E_1, R, E_2,\pi)\mapsto \big(m(E_1),\,R,\,m(E_2),\,\pi\big),
\]
and obtain the canonical set $\widehat{\mathcal{T}}_c=\{\Phi(t):t\in\mathcal{T}_c^{\mathrm{clean}}\}$.

\subsection{Probe Creation}
We created two types of Probes: the first built on $\mathbb{G}_3$ and the second built on the intersection of $\mathbb{G}_1$ and $\mathbb{G}_2$ $\rightarrow (\mathbb{G}_1 \cap \mathbb{G}_2)$

\subsubsection{Probe 1}
Containing 100 multiple-choice questions, this probe tests the joint AD+T2DM query. The questions are framed to cover single hop, multi-hop and fill in the blank (FITB).  Let $A\in\{0,1\}^{|\widehat{\mathcal{E}}_3|\times|\widehat{\mathcal{E}}_3|}$ be the adjacency over $\mathcal{R}_c$.
\begin{itemize}
  \item \textbf{Single-hop.} Choose $(u,r,v)\!\in\!\widehat{\mathcal{T}}_{c,3}$ and create one correct option and three distractors from $\mathcal{N}^{-}(u)=\{x:A_{x,u}=1\}$ that match type/frequency. This checks one clean link.
  \item \textbf{Multi-hop, pair-selection.} For a target $x$, the set of \emph{direct} causes is $\mathcal{P}_1(x)=\{u:A_{u,x}=1\}$. The correct answer is an unordered pair $\{u,v\}\subseteq\mathcal{P}_1(x)$. Distractors come from $\mathcal{P}_2(x)=\{u:(A^2)_{u,x}=1\}$ or close neighbors that look right but are not immediate parents. This stresses real 2-hop reasoning.
  \item \textbf{FITB.} Mask one canonical token in $(u,r,v)$; only the canonical label passes. This drills precision under synonyms.
\end{itemize}
We ensure synonym control explicitly, as options must be distinct in canonical space ($m(c_i)\neq m(c_j)$), so there are no duplicate answers under different spellings.

\begin{table*}[ht]
\centering
\caption{QA items formatted as fill-in-the-gap, multi-hop, and single-hop examples.}
\scriptsize
\setlength{\tabcolsep}{4pt}
\renewcommand{\arraystretch}{1.15}
\begin{adjustbox}{max width=\linewidth}
\begin{tabularx}{\linewidth}{
  >{\raggedright\arraybackslash}p{.16\linewidth}
  >{\raggedright\arraybackslash}p{.36\linewidth}
  >{\raggedright\arraybackslash}p{.22\linewidth}
  >{\raggedright\arraybackslash}p{.26\linewidth}}
\toprule
\textbf{Question type} & \textbf{Question} & \textbf{Options} & \textbf{Answer choices} \\
\midrule
Fill-in-the-gap &
Type 2 diabetes increases Alzheimer’s risk through \underline{\hspace{1.2cm}} and \underline{\hspace{1.2cm}}. &
\begin{tabular}[t]{@{}l@{}}
1. Neuroinflammation\\
2. Amyloid degradation\\
3. Insulin resistance\\
4. Tau dephosphorylation
\end{tabular} &
\begin{tabular}[t]{@{}l@{}}
A: 1 and 2\\
B: 3 and 4\\
C: 3 and 2\\
D: 2 and 4\\
E: 3 and 1
\end{tabular}
\\
\addlinespace
Multi-hop &
Which two are direct precursors of neuroinflammation escalation? &
\begin{tabular}[t]{@{}l@{}}
1. A$\beta$ oligomers\\
2. Peripheral infection\\
3. Aerobic fitness\\
4. Microglial priming
\end{tabular} &
\begin{tabular}[t]{@{}l@{}}
A: 1 and 2\\
B: 1 and 4\\
C: 2 and 3\\
D: 3 and 4
\end{tabular}
\\
\addlinespace
Single-hop &
Abnormal insulin signaling in the brain primarily results in: &
\begin{tabular}[t]{@{}l@{}}
A: Lower GSK3$\beta$ activity\\
B: Reduced oxidative damage\\
C: Enhanced mitochondrial function\\
D: Higher synaptic resilience\\
E: Reduced A$\beta$ degradation and increased tau phosphorylation
\end{tabular} &
— \\
\bottomrule
\end{tabularx}
\end{adjustbox}
\label{tab:qa_items}
\end{table*}

\subsubsection{Probe 2}
This probe targets $\mathbb{G}_1 \cap \mathbb{G}_2$. We embed the triple with an encoder $s(\cdot)\in\mathbb{R}^d$ and use cosine similarity to screen for intersection candidates $\mathcal{I}$
\begin{equation}
\label{eq:cos-thresh}
\mathcal{I} \;=\; \Big\{ (\tau_1,\tau_2)\in \widehat{\mathcal{T}}_{c,1}\!\times\!\widehat{\mathcal{T}}_{c,2}:\ \mathrm{cos}\!\left(s(\tau_1),s(\tau_2)\right)\ge 0.65 \Big\}.
\end{equation}
Yielding $|\mathcal{I}|=424$. After stop-word removal and de-duplication in canonical space, we keep $|\widehat{\mathcal{I}}|=193$ items. We then form questions from this intersection subgraph:
\begin{itemize}
  \item \textbf{Single-hop, FITB:} Same as Probe~1 but restricted to $\widehat{\mathcal{I}}$.
  \item \textbf{Multi-hop with direction.} If the true edge is $u\!\to\!x$, then the option $x\!\to\!u$ is wrong even though it uses the same names. We present four atomic options (1--4) and ask for the correct pair among A--E (the 2-combinations). Exactly two pairs are correct; both must point into the target.
\end{itemize}
0.65 was chosen as the cosine similarity value because it kept the same causal theme in both diseases (e.g., insulin signaling, inflammation) without having too loose matches.

\subsection{RAG setups Prompting and Answer Generation}
We compare six retrieval setups:
\[
K \in \big\{\varnothing\,,\ \mathbb{G}_1,\ \mathbb{G}_2,\ \mathbb{G}_1\!+\!\mathbb{G}_2,\ \mathbb{G}_3,\ \mathbb{G}_1\!+\!\mathbb{G}_2\!+\!\mathbb{G}_3\big\}.
\]

Where $\varnothing\ \rightarrow \text{(No-RAG)}$. For each value of $K$, we run the same LLM with and without this context. For QA we use $T\in\{0,0.2,0.5\}$; Lower $T$ sticks closer to retrieved facts; higher $T$ can add details but risks drift.
All the prompts were zero-shot instruction based. 
\begin{ashbox}
\ttfamily
You are answering a multiple-choice question.\\
Return ONLY one uppercase letter from this set: \{allowed\_str\}.\\
Do not include explanations or extra text.
\end{ashbox}

Since all questions were multiple-choice, we incorporated common distractors in the options. For questions with multiple answers, we tested the macro and micro F1 in those cases.

\section{Results}

\subsection{Effect of Varying Temperature}
For each Model$\times$Probe$\times$Graph configuration, we compared macro-F$_1$ across temperatures using pairwise Welch's $t$-tests (unequal variances) for the three temperature values: $0$ vs.\ $0.2$, $0.2$ vs.\ $0.5$, and $0$ vs.\ $0.5$. Within each configuration we applied Holm–Bonferroni correction across the three tests; statistical significance is denoted as * ($p_{\text{adj}}{<}.05$), ** ($p_{\text{adj}}{<}.01$), and *** ($p_{\text{adj}}{<}.001$). Effect sizes (Cohen’s $d$) were computed for all comparisons.

Because each temperature condition is represented by a single run per configuration (i.e., $n{=}1$ per temperature), inferential tests were of low statistical power and produced no adjusted $p$-values below $.05$; consequently, no cells received a significance marker after Holm correction. We therefore report directional patterns descriptively. Across all 84 configurations, increasing temperature from $0{\to}0.5$ reduced macro-F$_1$ in 52 cases, increased it in 23, and left it unchanged in 9 (median $\Delta=-0.02$). The attenuation with higher $T$ was more pronounced on Probe 1 (28 decreases, 9 increases; median $\Delta=-0.03$) than Probe 2 (24 decreases, 14 increases; median $\Delta=-0.01$). By graph condition, $\mathbb{G}_1$ was most temperature-stable (balanced 7 increases/7 decreases; median $\Delta\approx0$), whereas $\mathbb{G}_2$, $\mathbb{G}_3$, and $\mathbb{G}_1$+$\mathbb{G}_2$(+$\mathbb{G}_3$) skewed toward decreasing values, however, it was small (median $\Delta\in[-0.03,-0.015]$). By model, \texttt{Anthropic.Claude-3-Haiku} showed the mildest tendency to improve with temperature (median $\Delta{=}+0.005$), while \texttt{Mistral-7B-Instruct-v0.3} exhibited the largest typical drop (median $\Delta{=}-0.12$). Illustratively, the largest decrease occurred for \texttt{Mistral-7B-Instruct-v0.3} on Probe~1 across several graphs ($\Delta_{0{\to}0.5}\in[-0.25,-0.21]$), whereas a notable increase was observed for \texttt{Mixtral-8x7B-v0.1} (No-RAG, Probe 2; $\Delta_{0{\to}0.5}{=}+0.09$).

On the domain of single-run estimations (no stars beyond the Holm adjustment), on the graphs of macro-F$_1$, the trend of higher decoding temperature is, in most cases, downward, particularly Probe~1 and non-$\mathbb{G}_1$ graphs. When stability is preferred, $T{=}0$ is the safest default; small exploration at $T{=}0.2$ may be helpful in some environments, but $T{=}0.5$ tends to worsen the accuracy. Accordingly, we take an averaged value of each of these temperatures.

\begin{table}[t]
\centering
\scriptsize
\caption{Temperature sensitivity by graph (macro-F$_1$). Counts across all Model$\times$Probe settings; $\Delta$ is median change from $T{=}0$ to $0.5$.}
\begin{tabular}{lrrrr}
\hline
Graph & \# Increases & \# Decreases & \# No change & Median $\Delta$ (0$\to$0.5) \\
\hline
$\mathbb{G}_1$         & 7  & 7  & 0 &  0.00 \\
$\mathbb{G}_1$+$\mathbb{G}_2$      & 3  & 10 & 1 & -0.03 \\
$\mathbb{G}_1$+$\mathbb{G}_2$+$\mathbb{G}_3$   & 3  & 9  & 2 & -0.03 \\
$\mathbb{G}_2$         & 3  & 10 & 1 & -0.03 \\
$\mathbb{G}_3$         & 3  & 8  & 3 & -0.02 \\
No-RAG     & 4  & 8  & 2 & -0.02 \\
\hline
\end{tabular}
\label{tab:temp-by-graph}
\end{table}


\begin{table*}[ht] 
\centering 
\caption{Results of Probe 1, Averaged across the three temperatures (0.0, 0.2 and 0.5)}
\label{tab:probe1} 
\scriptsize 
\begin{tabular}{l l c c c c c c}
\hline
\multicolumn{2}{c}{} & \multicolumn{6}{c}{Systems} \\
\cline{3-8}
Model & Metric & $\mathbb{G}_1$ & $\mathbb{G}_1$+$\mathbb{G}_2$ & $\mathbb{G}_1$+$\mathbb{G}_2$+$\mathbb{G}_3$ & $\mathbb{G}_2$ & $\mathbb{G}_3$ & No-RAG \\
\hline
\multirow{2}{*}{Anthropic.Claude-3-Haiku} & Acc/ F$_{1Micro}$ & 0.84 / 0.84 & \textbf{0.92} / \textbf{0.92} & \textbf{0.93} / \textbf{0.93} & \textbf{0.93} / \textbf{0.93} & 0.89 / 0.89 & \textbf{0.96} / \textbf{0.96} \\

& Macro P/R/F$_1$ & 0.87 / 0.84 / 0.84 & \textbf{0.93} / \textbf{0.93} / \textbf{0.93} & \textbf{0.94} / \textbf{0.93} / \textbf{0.93} & \textbf{0.94} / \textbf{0.94} / \textbf{0.93} & 0.91 / 0.90 / 0.90 & 0.91 / 0.91 / 0.91 \\
\hline
\multirow{2}{*}{GPT-OSS-20B} & Acc/ F$_{1Micro}$ & \textbf{0.87} / \textbf{0.87} & 0.85 / 0.85 & 0.82 / 0.82 & 0.82 / 0.82 & 0.84 / 0.84 & 0.91 / 0.91 \\
& Macro P/R/F$_1$ & \textbf{0.88} / \textbf{0.87} / \textbf{0.87} & 0.87 / 0.86 / 0.86 & 0.85 / 0.83 / 0.82 & 0.84 / 0.82 / 0.82 & 0.87 / 0.85 / 0.85 & 0.92 / 0.92 / 0.91 \\
\hline
\multirow{2}{*}{Llama-3.1-8B-Instruct} & Acc/ F$_{1Micro}$ & 0.69 / 0.69 & 0.63 / 0.63 & 0.65 / 0.65 & 0.66 / 0.66 & 0.67 / 0.67 & 0.83 / 0.83 \\
& Macro P/R/F$_1$ & 0.76 / 0.66 / 0.67 & 0.69 / 0.62 / 0.62 & 0.74 / 0.63 / 0.65 & 0.71 / 0.65 / 0.65 & 0.74 / 0.65 / 0.66 & 0.85 / 0.82 / 0.83 \\
\hline
\multirow{2}{*}{Llama-3.3-70B-Instruct} & Acc/ F$_{1Micro}$ & 0.86 / 0.86 & 0.91 / 0.91 & 0.92 / 0.92 & 0.91 / 0.91 & \textbf{0.92} / \textbf{0.92} & 0.95 / 0.95 \\
& Macro P/R/F$_1$ & 0.72 / 0.71 / 0.71 & 0.91 / 0.91 / 0.91 & 0.93 / 0.92 / 0.92 & 0.91 / 0.90 / 0.90 & \textbf{0.92} / \textbf{0.92} / \textbf{0.92} & \textbf{0.96} / \textbf{0.96} / \textbf{0.96} \\
\hline
\multirow{2}{*}{Mistral-7B-Instruct-v0.3} & Acc/ F$_{1Micro}$ & 0.67 / 0.67 & 0.68 / 0.68 & 0.69 / 0.69 & 0.73 / 0.73 & 0.72 / 0.72 & 0.67 / 0.67 \\
& Macro P/R/F$_1$ & 0.79 / 0.67 / 0.70 & 0.75 / 0.69 / 0.70 & 0.77 / 0.70 / 0.71 & 0.78 / 0.73 / 0.74 & 0.78 / 0.72 / 0.73 & 0.78 / 0.66 / 0.69 \\
\hline
\multirow{2}{*}{Mixtral-8x7B-v0.1} & Acc/ F$_{1Micro}$ & 0.83 / 0.83 & 0.83 / 0.83 & 0.87 / 0.87 & 0.88 / 0.88 & 0.84 / 0.84 & 0.79 / 0.79 \\
& Macro P/R/F$_1$ & 0.86 / 0.83 / 0.84 & 0.82 / 0.80 / 0.80 & 0.88 / 0.87 / 0.87 & 0.90 / 0.89 / 0.89 & 0.86 / 0.85 / 0.84 & 0.85 / 0.79 / 0.80 \\
\hline
\multirow{2}{*}{Qwen2.5-32B-Instruct} & Acc/ F$_{1Micro}$ & 0.85 / 0.85 & 0.90 / 0.90 & 0.89 / 0.89 & 0.89 / 0.89 & 0.87 / 0.87 & 0.91 / 0.91 \\
& Macro P/R/F$_1$ & 0.84 / 0.84 / 0.84 & 0.90 / 0.89 / 0.89 & 0.89 / 0.88 / 0.88 & 0.89 / 0.87 / 0.88 & 0.87 / 0.87 / 0.87 & 0.92 / 0.90 / 0.91 \\
\hline
\end{tabular}
\end{table*}

\begin{table*}[ht] 
\centering 
\caption{Results of Probe 2, Averaged across the three temperatures (0.0, 0.2 and 0.5)}
\label{tab:probe2} 
\scriptsize 
\begin{tabular}{l l c c c c c c}
\hline
\multicolumn{2}{c}{} & \multicolumn{6}{c}{Systems} \\
\cline{3-8}
Model & Metric & $\mathbb{G}_1$ & $\mathbb{G}_1$+$\mathbb{G}_2$ & $\mathbb{G}_1$+$\mathbb{G}_2$+$\mathbb{G}_3$ & $\mathbb{G}_2$ & $\mathbb{G}_3$ & No-RAG \\
\hline
\multirow{2}{*}{Anthropic.Claude-3-Haiku} & Acc/ F$_{1Micro}$ & \textbf{0.58} / \textbf{0.58} & \textbf{0.60} / \textbf{0.60} & \textbf{0.62} / \textbf{0.62} & 0.61 / 0.61 & 0.53 / 0.53 & 0.61 / 0.61 \\
& Macro P/R/F1 & 0.53 / 0.52 / 0.50 & 0.57 / 0.57 / \textbf{0.55} & \textbf{0.60} / \textbf{0.60} / \textbf{0.57} & 0.56 / 0.55 / 0.54 & 0.50 / 0.49 / 0.47 & 0.54 / 0.55 / 0.55 \\
\hline
\multirow{2}{*}{GPT-OSS-20B} & Acc/ F$_{1Micro}$ & 0.46 / 0.46 & 0.50 / 0.50 & 0.48 / 0.48 & 0.50 / 0.50 & 0.44 / 0.44 & 0.43 / 0.43 \\
& Macro P/R/F1 & 0.56 / 0.45 / 0.45 & 0.56 / 0.47 / 0.46 & 0.57 / 0.47 / 0.46 & \textbf{0.63} / 0.51 / 0.48 & \textbf{0.54} / 0.46 / 0.41 & 0.49 / 0.42 / 0.39 \\
\hline
\multirow{2}{*}{Llama-3.1-8B-Instruct} & Acc/ F$_{1Micro}$ & 0.53 / 0.53 & 0.51 / 0.51 & 0.48 / 0.48 & 0.56 / 0.56 & \textbf{0.54} / \textbf{0.54} & 0.61 / 0.61 \\
& Macro P/R/F1 & 0.44 / 0.41 / 0.41 & 0.47 / 0.41 / 0.42 & 0.40 / 0.38 / 0.37 & 0.52 / 0.44 / 0.45 & 0.45 / 0.41 / 0.40 & 0.51 / 0.50 / 0.50 \\
\hline
\multirow{2}{*}{Llama-3.3-70B-Instruct} & Acc/ F$_{1Micro}$ & \textbf{0.58} / \textbf{0.58} & 0.54 / 0.54 & 0.54 / 0.54 & 0.59 / 0.59 & 0.52 / 0.52 & 0.56 / 0.56 \\
& Macro P/R/F1 & 0.52 / 0.51 / 0.51 & 0.47 / 0.47 / 0.47 & 0.51 / 0.50 / 0.49 & 0.53 / 0.52 / 0.52 & 0.48 / 0.49 / \textbf{0.48} & 0.52 / 0.49 / 0.49 \\
\hline
\multirow{2}{*}{Mistral-7B-Instruct-v0.3} & Acc/ F$_{1Micro}$ & 0.36 / 0.36 & 0.40 / 0.40 & 0.39 / 0.39 & 0.44 / 0.44 & 0.35 / 0.35 & 0.38 / 0.38 \\
& Macro P/R/F1 & 0.32 / 0.26 / 0.23 & 0.46 / 0.31 / 0.29 & 0.36 / 0.28 / 0.25 & 0.46 / 0.34 / 0.33 & 0.38 / 0.27 / 0.24 & 0.40 / 0.31 / 0.29 \\
\hline
\multirow{2}{*}{Mixtral-8x7B-v0.1} & Acc/ F$_{1Micro}$ & 0.47 / 0.47 & 0.58 / 0.58 & 0.57 / 0.57 & 0.56 / 0.56 & 0.47 / 0.47 & 0.46 / 0.46 \\
& Macro P/R/F1 & 0.44 / 0.43 / 0.42 & 0.55 / 0.54 / 0.54 & 0.51 / 0.50 / 0.50 & 0.52 / 0.51 / 0.51 & 0.45 / 0.44 / 0.43 & 0.47 / 0.39 / 0.39 \\
\hline
\multirow{2}{*}{Qwen2.5-32B-Instruct} & Acc/ F$_{1Micro}$ & 0.56 / 0.56 & 0.56 / 0.56 & 0.55 / 0.55 & \textbf{0.64} / \textbf{0.64} & 0.51 / 0.51 & \textbf{0.65} / \textbf{0.65} \\
& Macro P/R/F1 & \textbf{0.60} / \textbf{0.60} / \textbf{0.54} & \textbf{0.59} / \textbf{0.61} / \textbf{0.55} & 0.53 / 0.55 / 0.52 & 0.62 / \textbf{0.65} / \textbf{0.61} & 0.51 / \textbf{0.53} / \textbf{0.48} & \textbf{0.61} / \textbf{0.63} / \textbf{0.60} \\
\hline
\end{tabular}
\end{table*}

\begin{figure*}[ht]
  \centering
  \begin{subfigure}{0.49\linewidth}
    \centering
    \includegraphics[width=\linewidth]{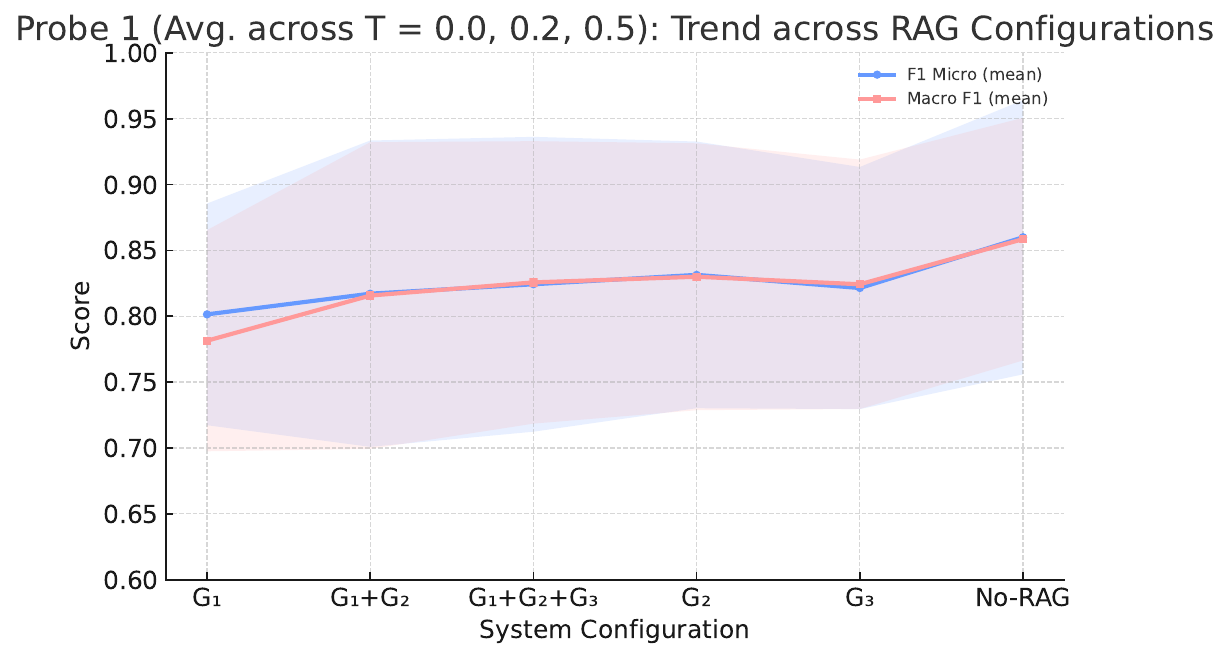}
    \caption{Probe 1}
    \label{fig:probe1}
  \end{subfigure}\hfill
  \begin{subfigure}{0.49\linewidth}
    \centering
    \includegraphics[width=\linewidth]{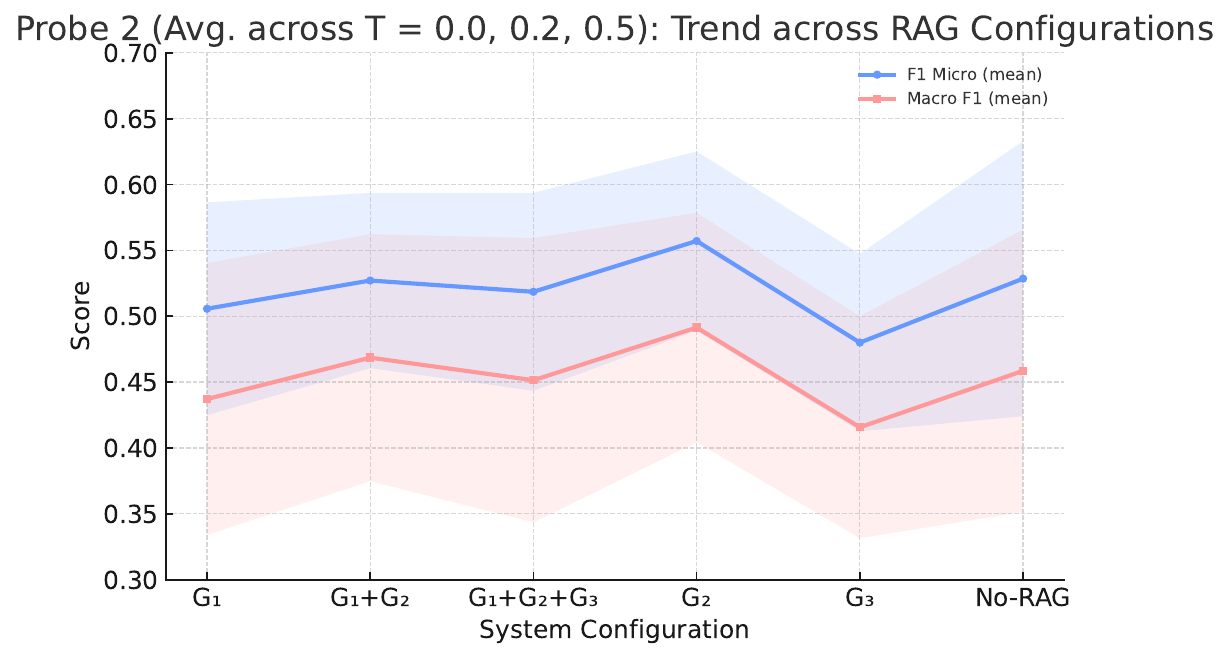}
    \caption{Probe 2}
    \label{fig:probe2}
  \end{subfigure}
  \caption{Trends across RAG configurations for the averaged F$_1$ scores of the models in Table \ref{tab:probe1} and \ref{tab:probe2} (averaged over $T=\{0.0,0.2,0.5\}$).}
  \label{fig:probe-trends}
\end{figure*}

\subsection{Model Performance}
The models which perform well on both probes are: Anthropic Claude-3-Haiku, (which is the only non-open source model), Qwen-2.5-32B-Instruct, Llama-3.1-8B-Instruct, Llama-3.3-70B-Instruct, and GPT-OSS-20B. Llama-3.3-70B-Instruct scores the highest macro F1 on the probe 1, while Qwen-2.5-32B-Instruct has the highest macro F1 in probe 2, beating the Anthropic Claude-3-Haiku model. An observation is that these models which are particularly large in size are already trained to contain large amounts of biomedical knowledge and therefore adding external information in the shape of a domain-specific KG is likely to have very minimal value and can even cause a model to become confused and confused by irrelevant or contradictory information, shown by the dip in the results, in Table \ref{tab:probe1} and \ref{tab:probe2}. From the results, we see that normal RAG might lure in erroneous or conflicting facts, which is likely the reason why $\mathbb{G}_1$ and $\mathbb{G}_3$ fell in performance in these models. On probe 2, the image is a bit different - RAG occasionally helps. For example, the combined graph ($\mathbb{G}_1$ + $\mathbb{G}_2$ + $\mathbb{G}_3$) slightly improves Anthropic Haiku from 0.61 to 0.63 and GPT-OSS-20B from 0.46 to 0.57, suggesting that even large models can benefit from external knowledge when tackling more complex, multi-hop reasoning tasks.

Small models benefit from $\mathbb{G}_2$. The baseline scores of mistral-7B and Mixtral-8X7B are moderate (0.69 and 0.80) in probe 1 and (0.29 and 0.39) in probe 1. The performance of both probes is significantly improved when they retrieve out of the AD-oriented KG ($\mathbb{G}_2$). On probe 1 for Mixtral-8X7B, F$_1$ is increased by 0.80 to 0.89, and probe 2 F$_1$ is also increased by 0.39 to 0.51. Mistral-7B exhibited a similar behavior, from 0.69 to 0.74 for probe 1 and 0.29 to 0.33 in probe 2. Interestingly, $\mathbb{G}_1$ + $\mathbb{G}_2$ is slightly worse than $\mathbb{G}_2$ alone, indicating that the most valuable information in these models is the AD graph instead of being part of the union with T2DM knowledge. The above improvements suggest that smaller models that possess less built-in biomedical knowledge can benefit meaningfully in terms of structured biomedical knowledge to fill in the gaps in their parameter knowledge. 

Performance in $\mathbb{G}_3$ showed a downward spike, even though there was no conclusive evidence as to why. In the models, except in Mistral and Mixtral, the $\mathbb{G}_3$ graph reduces the F$_1$. An interesting case of the Llama-3.1-8B  Instruct model shows that the combined larger graph: $\mathbb{G}_1$ + $\mathbb{G}_2$ + $\mathbb{G}_3$ dropped the performance from 0.85 to 0.65 on probe 1. 0.65 was the same value as when using $\mathbb{G}_2$ alone, and was even less (0.62) on $\mathbb{G}_1$ alone.  Probe 2 was quite similar, as we saw a drop in the combined graph of  $\mathbb{G}_1$ + $\mathbb{G}_2$ + $\mathbb{G}_3$ fom 0.50 to 0.40 macro F$_1$. We hypothesized that a larger KG may add irrelevant relations, and our findings validate this, meaning that the broader graph may mislead LLMs, exposing them to unfounded relations. 

Upon adding up the improvement over the No-RAG baseline across all the models and all the probes, one can easily see that $\mathbb{G}_2$ is positive (mean improvement +0.006 F$_1$) and $\mathbb{G}_1$, $\mathbb{G}_3$ and the combination of graphs have negative mean improvements (–0.04 to -0.01). Therefore, the graph on AD ($\mathbb{G}_2$) is the only domain-specific KG that was helpful.

\section{Discussion}
\label{sec:discussion}
It is experimentally demonstrated that the source of the retrieval and the scope of retrieval $\mathbb{G}_1$, $\mathbb{G}_2$, $\mathbb{G}_3$, and their combinations directly influence the quality of answers. Since Probe 1 is derived out of $\mathbb{G}_3$, the resulting merged KG is likely to include the accurate facts sought by a large number of questions; whereas Probe 2 is created out of $\mathbb{G}_1 \cap \mathbb{G}_2$, thus signals that are similar across the two domains of diseases have a stronger influence. In both probes, the addition of additional sources ($\mathbb{G}_1$, $\mathbb{G}_2$, $\mathbb{G}_3$) does not assure higher accuracy: the addition of breadth raises recall but may introduce many other unrelated distracting facts that reduce precision.

\newcommand{\sym}[1]{\ifmmode^{#1}\else\(^{#1}\)\fi}

\begin{table*}[t]
  \centering
  \small
    \caption{RAG system comparison on {Probe 1}: Macro-F1 and significance vs.\ No-RAG}
    \label{tab:rag_probe1}
    \begin{tabular}{lcccccc}
      \toprule
      \textbf{Model} & \textbf{No-RAG} & $\mathbb{G}_1$ & $\mathbb{G}_2$ & $\mathbb{G}_1$+$\mathbb{G}_2$ & $\mathbb{G}_3$ & $\mathbb{G}_1$+$\mathbb{G}_2$+$\mathbb{G}_3$ \\
      \midrule
      Llama-3.1-8B-Instruct    & 0.83 & 0.67\sym{***} & 0.65\sym{**}  & 0.62\sym{***} & 0.66\sym{***} & 0.65 \\
      Mistral-7B-Instruct-v0.3 & 0.69 & 0.70           & 0.74          & 0.70           & 0.73           & 0.71 \\
      Mixtral-8$\times$7B-v0.1 & 0.80 & 0.84           & 0.89\sym{**}  & 0.80           & 0.84           & 0.87\sym{*} \\
      Qwen-2.5-32B-Instruct    & 0.91 & 0.84\sym{*}    & 0.88          & 0.89           & 0.87\sym{*}    & 0.88 \\
      GPT-OSS-20B              & 0.91 & 0.87           & 0.82\sym{*}   & 0.86\sym{*}    & 0.85\sym{*}    & 0.82 \\
      Anthropic Claude-3-Haiku & 0.91 & 0.84           & 0.93          & 0.93           & 0.90           & 0.93 \\
      Llama-3.3-70B-Instruct   & 0.96 & 0.71\sym{***}  & 0.90\sym{**}  & 0.91\sym{*}    & 0.92\sym{*}    & 0.92\sym{**} \\
      \bottomrule
    \end{tabular}

      \footnotesize
      Notes: Stars denote Welch two-sample $t$-test vs.\ No-RAG using the three replicates per condition:
      \;$\sym{*}\,p{<}.05$, \;$\sym{**}\,p{<}.01$, \;$\sym{***}\,p{<}.001$.

\end{table*}

\begin{table*}[t]
  \centering
  \small
    \caption{RAG system comparison on {Probe 2}: Macro-F1  and significance vs.\ No-RAG}
    \label{tab:rag_probe2}
    \begin{tabular}{lcccccc}
      \toprule
      \textbf{Model} & \textbf{No-RAG} & $\mathbb{G}_1$ & $\mathbb{G}_2$ & $\mathbb{G}_1$+$\mathbb{G}_2$ & $\mathbb{G}_3$ & $\mathbb{G}_1$+$\mathbb{G}_2$+$\mathbb{G}_3$     \\
      \midrule
      Llama-3.1-8B-Instruct    & 0.50 & 0.41\sym{*}  & 0.45         & 0.42\sym{*}    & 0.40\sym{*}  & 0.37\sym{**} \\
      Mistral-7B-Instruct-v0.3 & 0.29 & 0.23\sym{*}  & 0.33         & 0.29           & 0.24\sym{*}  & 0.25 \\
      Mixtral-8$\times$7B-v0.1 & 0.39 & 0.42         & 0.51\sym{*}  & 0.54\sym{*}    & 0.43         & 0.50\sym{*} \\
      Qwen-2.5-32B-Instruct    & 0.60 & 0.54\sym{**} & 0.61         & 0.55\sym{*}    & 0.48\sym{**} & 0.52\sym{**} \\
      GPT-OSS-Bb              & 0.39 & 0.45         & 0.48\sym{*}  & 0.46           & 0.41         & 0.46 \\
      Anthropic Claude-3-Haiku & 0.55 & 0.50\sym{*}  & 0.54         & 0.55           & 0.47\sym{**} & 0.57\sym{*} \\
      Llama-3.3-70B-Instruct   & 0.49 & 0.51         & 0.52         & 0.47           & 0.48         & 0.49 \\
      \bottomrule
    \end{tabular}
      \footnotesize
     
     Notes: Stars denote Welch two-sample $t$-test vs.\ No-RAG using the three replicates per condition:
      \;$\sym{*}\,p{<}.05$, \;$\sym{**}\,p{<}.01$, \;$\sym{***}\,p{<}.001$.
\end{table*}

\subsection{Significance}
We provide statistical significance of each condition (three independent runs each model-probe-system) (paired with paired two-sample $t$-tests against No-RAG): $^{*}\!p<0.05$, $^{**}\!p<0.01$, $^{***}\!p<0.001$. Table \ref{tab:rag_probe1} and \ref{tab:rag_probe2} only indicate differences that are of significance at these values; non-stars would not be statistically different to the baseline at $\alpha{=}0.05$.

\subsection{How RAG scope interacts with each probe}

\paragraph{Probe 1}
This probe was constructed out of direct existing knowledge, and systems that used $\mathbb{G}_3$ directly, accessed a single graph that happened to answer all Probe 1 facts; however, these did not help these systems, but rather introduced extraneous context, which caused failure:
\begin{itemize}
  \item Large generalist models (Llama-3.3-70B). Adding additional, tangential evidence, on the basis of $\mathbb{G}_1$ only, generated a huge, significant decline ($^{***}$), smaller, but significant, declines in the case of $\mathbb{G}_2$, $\mathbb{G}_3$, and $\mathbb{G}_1$+$\mathbb{G}_2$ ($^{*}$–$^{**}$). The model already captures a lot of knowledge that is required to respond to health-based questions.
  \item Mid-sized mixture models (Mixtral-8$\times$7B). $\mathbb{G}_2$ yields a \emph{significant} gain ($^{**}$) and $\mathbb{G}_1$+$\mathbb{G}_2$+$\mathbb{G}_3$ a smaller gain ($^{*}$), suggesting that well-scoped AD evidence helps when merged facts are relevant but not memorized.
  \item Smaller instruction models (Llama-3.1-8B). Against all KG-RAG settings, the performance is significantly low. One-sample t-test incorrect on Probe 1 ($^{*}$–$^{***}$): Sensitive to distractors in the case of retrieval of multiple semantically related facts.
  \item Other models. Mistral-7B shows no reliable change on Probe 1 (mostly unstarred), while Qwen-2.5-32B shows small but significant drops for $\mathbb{G}_1$ and $\mathbb{G}_3$ ($^{*}$).
\end{itemize}

\textbf{Why this pattern?}
Probe 1 questions reflect the merged space in existing literature, (AD\,+\,T2DM) i.e. different from just naively joining $\mathbb{G}_1$ and  $\mathbb{G}_2$. When retrieval comes from a narrower domain ($\mathbb{G}_1$ or $\mathbb{G}_2$) or an over-broad union ($\mathbb{G}_1$+$\mathbb{G}_2$+$\mathbb{G}_3$) the context either {misses} key merged relations or {dilutes} them with near-miss facts (e.g., disease-specific mechanisms that are irrelevant to the asked relation). High level models, with their substantial common-sense and medical knowledge inherent in them, can be readily diverted by near-misses. On the contrary, mid-sized models take advantage of focused AD signal ($\mathbb{G}_2$) when it coincides with merged facts of a probe.

\paragraph{Probe 2 (formulated on $\mathbb{G}_1\cap \mathbb{G}_2$).}
In this case, consistent evidence between the two diseases will be of greatest importance:
\begin{itemize}
  \item Mixtral-8$\times$7B. All of $\mathbb{G}_2$, $\mathbb{G}_1$+$\mathbb{G}_2$, $\mathbb{G}_1$+$\mathbb{G}_2$+$\mathbb{G}_3$ are improving significantly ($^{*}$), a fact that implies that AD-centric cues and their combination with T2DM are quite consistent with the intersection facts.
  \item Qwen-2.5-32B. The several settings of RAG ($\mathbb{G}_1$, $\mathbb{G}_1$+$\mathbb{G}_2$, $\mathbb{G}_3$, $\mathbb{G}_1$+$\mathbb{G}_2$+$\mathbb{G}_3$) result in important drops ($^{*}$–$^{**}$) and $\mathbb{G}_2$ is not statistically different to baseline. This indicates accuracy rather than breadth: AD-specific retrieval is less risky than the domain mixture of this probe.
  \item Mistral-7B. $\mathbb{G}_1$ and $\mathbb{G}_3$ indicate minor yet significant declines ($^{*}$); other settings are not significant.
  \item Llama-3.1-8B. The vast majority of KG-RAG environments impair performance ($^{*}$–$^{**}$), which once more indicates susceptibility to distractors in smaller models.
  \item GPT-OSS-20B \& Claude-Haiku. Mixed outcomes: some modest gains (e.g., Haiku with $\mathbb{G}_1$+$\mathbb{G}_2$+$\mathbb{G}_3$, $^{*}$) and several significant drops where the added context conflicts with the intersection signal.
  \item Llama-3.3-70B. None of the significant differences is observed- in keeping with a strong prior which is neither aided nor injured by the retrieved snippets.
\end{itemize}

\subsection{Explaining the patterns of stars}
We want to consider if the RAG is useful or harmful. There are three typical processes that we see:
\begin{enumerate}
  \item  Models are more potential to assist in the event that the construction space of the probe equals the retrieval space (Probe 1 with $\mathbb{G}_3$, Probe 2 with $\mathbb{G}_2$ or $\mathbb{G}_1$+$\mathbb{G}_2$). False positive (e.g. Probe 2 and $\mathbb{G}_3$) enhances contradictions and off-target cues and produces $^{*}$/$^{**}$ drops.
  \item  The recall is augmented by unions of ($\mathbb{G}_1$+$\mathbb{G}_2$), $\mathbb{G}_1$+$\mathbb{G}_2$)+$\mathbb{G}_3$) but it bring heterogeneous evidence to it. Otherwise, the models will overfittingly adapt to spurious but fluent responses, and will produce large drops even with truthful facts.
  \item Larger models tolerate imperfect retrieval better, (as we see in Table \ref{tab:rag_probe1} and \ref{tab:rag_probe2} where we have often unstarred or mixed effects. However, these models are also susceptible to {highly plausible} distractors; smaller models rely more heavily on retrieved text and thus amplify retrieval noise, leading to frequent $^{*}$–$^{***}$ drops.
\end{enumerate}

\subsection{Temperature effects}
System-to-system changes in temperature (0, 0.2, 0.5) had little influence on the conclusions of the RAG type: most in-system patterns were statistically insignificant, and significant ones were small compato system differences. In practice, these probes are mainly tuned by RAG choice; temperature tuning must always come second after a well-scoped graph is chosen and a strong retriever is selected.

\subsection{Frontier LLMs and Non-expert Baseline}
\label{subsec:frontier_llm_baseline}

We also posed both probes to two general-purpose LLMs and a non-expert human being. 

\begin{table}[t]
  \centering
  \scriptsize
    \caption{Gemini, ChatGPT, and non-expert human performance on Probe~1  and Probe~2.}
    \label{tab:frontier_llm_human}
    \begin{tabular}{lcc}
      \toprule
      \textbf{Agent} & \textbf{Probe 1 (100)} & \textbf{Probe 2 (110)} \\
      \midrule
      Gemini 2.5 Pro & 99/100 (99\%) & 69/110 (62.7\%) \\
      ChatGPT 5 Thinking & 98/100 (98\%) & 77/110 (70.0\%) \\
      Naive Human (no medical knowledge) & 38/100 (38\%) & 30/110 (27.3\%) \\
      \bottomrule
    \end{tabular}
\end{table}

Table~\ref{tab:frontier_llm_human} contrasts the accuracy of Gemini, ChatGPT, and a non-expert human across Probe~1 and Probe~2. Both frontier models achieve near-ceiling performance on Probe~1 (99\% and 98\%, respectively), revealing that the initial probe tasks are largely saturated. In contrast, Probe~2 introduces greater conceptual and retrieval difficulty, producing a notable accuracy decline—Gemini drops by 36.3\% points, while ChatGPT falls by 28\%. The naive human baseline shows only a modest relative decline, but its overall accuracy remains far below that of the models. The naive human baseline is a test for random guessing, as the people who took the test did not have prior knowledge of the field and were encouraged to guess randomly.

When compared with the open-weight counterparts in Tables~\ref{tab:probe1} and~\ref{tab:probe2}, the frontier models display markedly higher stability and consistency. Anthropic Claude-3-Haiku, the strongest among the smaller systems, achieved accuracies around 96\% on Probe 1 but fell to roughly 61\% on Probe 2, a pattern mirrored by all the other models. This shows that Probe~2 acts as a discriminative benchmark, separating models with generalizable reasoning (ChatGPT and Claude-tier systems) from those whose performance is more retrieval-anchored or domain-fragile.

ChatGPT’s smaller degradation indicates stronger robustness to task complexity and class imbalance, aligning with its narrower gap between Micro and Macro F$_1$ in Probe~2. Gemini’s sharper fall suggests sensitivity to rare or ambiguous cases, reinforcing that Probe~2 better exposes differential reasoning depth rather than surface-level recall. Overall, frontier LLMs far exceed human baselines, yet their relative separation on Probe~2 highlights the probe’s discriminative power for evaluating balanced generalization.

\subsection{Error Analysis}
We covered the following errors shown in Table \ref{tab:error-breakdown}: (1) {Directionality flips} (picked $B\!\to\!A$ instead of $A\!\to\!B$), (2) {Two-hop chain order errors} (mis-ordered $A\!\to\!B\!\to\!C$ pairs), (3) {Negation/exception misreads} (e.g., ``\emph{is not associated}'' / ``\emph{pick the exception}''), 
(4) {Immediate vs.\ downstream cause} (selected a true but non-proximal effect),
(5) {Undefined token guesswork} (e.g., ambiguous labels like ``FX protein'') and
(6) {Overweighting canonical AD triad} (A$\beta$/tau/neuroinflammation selected when vascular/metabolic was targeted)

\begin{table}[t]
\scriptsize
\centering
\caption{Miss classification on Probe~2 (33 total errors).}
\label{tab:error-breakdown}
\begin{tabular}{lrr}
\toprule
\textbf{Category} & \textbf{Count} & \textbf{Percent} \\
\midrule
Directionality flips & 9 & 27.27\% \\
Two-hop chain order & 8 & 24.24\% \\
Negation/exception & 6 & 18.18\% \\
Immediate vs.\ downstream & 5 & 15.15\% \\
Undefined token guesswork & 3 & 9.09\% \\
AD-triad overweighting & 2 & 6.06\% \\
\midrule
\textbf{Total} & \textbf{33} & \textbf{100\%} \\
\bottomrule
\end{tabular}
\end{table}

\subsection{Recommendations}
\begin{itemize}
  \item When adding several graphs, do so conditionally: by putting in a ranker that favors off-topic passages being demoted and evidence being given precedence that is found in both domains of the disease.
  \item Smaller models benefit most from clean, highly relevant passages, while larger models need less context overall but require stricter filtering to avoid distractors.
  \item Given three runs per condition, using star-coding ($^{*}$/$^{**}$/$^{***}$) is essential to avoid over-interpreting small differences in mean values that might not be statistically meaningful.
\end{itemize}

\paragraph{Threats to validity}
(1) Only three replicates per cell limit power; results marked unstarred could still harbor small effects. (2)  Factual mix reproducible by our KG construction decisions (entity normalization, relation filtering, and co-reference backbone) determines the factual mix that can be accessed, and any improvement or error in this regard directly translates into the results of QA. (3) Prompting and re-ranking were held fixed; stronger retrieval/re-ranking may change the balance between precision and recall.

\section{Conclusion}

This work studied domain-specific KG-RAG for healthcare LLMs under realistic design choices: graph scope ($\mathbb{G}_1$/$\mathbb{G}_2$/$\mathbb{G}_3$), probe definition (merged vs.\ intersection), model capacity, and decoding temperature. Three consistent lessons emerged.

\textbf{(1) Match retrieval scope to task scope.} Probe 1 (merged AD -T2DM relations) is best served by retrieval that is to a large extent coincidental with the relations (e.g. $\mathbb{G}_3$ or more specific union of $\mathbb{G}_2$), whereas Probe 2 (intersection-style questions) is best served by $\mathbb{G}_2$ or $\mathbb{G}_1$+$\mathbb{G}_2$ rather than excessively broad unions.

\textbf{(2) Favor precision over breadth.} Uniting graphs increases recall but also injects heterogeneous evidence; without strong ranking/filters, distractors lower accuracy. Precision-first retrieval with scope-matched graphs is more reliable.

\textbf{(3) Right-size KG-RAG to model capacity.} Smaller and mid-sized models gain the most from clean, well-scoped retrieval (notably with $\mathbb{G}_2$). Larger models often match or surpass KG-RAG with \textit{No-RAG} on merged-scope questions, reflecting strong parametric knowledge and a higher sensitivity to noisy context.

 Practically, teams should: (i) choose graphs that match their question distribution; (ii) deploy rankers/filters that shed off-topic spans and reward corroborated evidence; and (iii) tailor retrieval strictness to model size. Future work includes risk-aware reranking, dynamic graph selection conditioned on query intent, and multi-hop reasoning that exploits KG structure without flooding prompts with near-miss facts.

\section{Limitations}
\label{sec:limitations}
\noindent We highlight key limitations and threats to validity of this study.

\begin{enumerate}
    \item Our graphs are centered on T2DM ($\mathbb{G}_1$), AD ($\mathbb{G}_2$) and AD+T2DM ($\mathbb{G}_3$) space. Findings might not be related to other conditions, specialities, non-PubMed corpora, and multilingual environments

    \item Although co-reference and canonicalization have been improved, the errors in entity connecting, synonym merging, and relation extracting may be carried over to the retrieval to produce plausible yet off-target evidence that diminishes accuracy. We used rule-based filtering to reduce false positives, but the graph is not 100\% perfect.

    \item We did not actively search retriever depth, hybrid lexical neural retrieval or learning-to-rank rerankers. A more powerful ranking stack that has the potential to minimize distractors, particularly on union graphs.

    \item Accuracy, Macro P/R/F1, and Micro F1 fail to reflect on calibration, factual grounding to primary sources and clinical harm potential. The fact-checking based on human judgment and reference was out of the question.

    \item The findings on smaller vs. larger models might not be generalizable to other architectures, tokenizer selection, alignment processes or domain-trained checkpoints.

    \item Our setup omits latency, cost, privacy, and PHI-handling constraints crucial in clinical workflows in this pilot study. Deployment-time retrieval drift, updates to literature, and governance requirements are not modeled.

    \item Since we had to run on a few resources, we did not carry out large multi-seed reruns, ablations of KG preprocessing steps, or confidence-interval reporting over all settings; here, some of the effects may be due to stochasticity.
\end{enumerate}

\small
\ifCLASSOPTIONcompsoc
  \section*{Acknowledgments}
\else
  \section*{Acknowledgment}
\fi

The authors would like to thank Paul Josiah, Toluwalase Kunle-John and Elizabeth Oyegoke for volunteering to take the test for the naive human baseline.



%



\bibliographystyle{IEEEtran}
\bibliography{references}  
\end{document}